# ASMDD: Arabic Speech Mispronunciation Detection Dataset

Salah A. Aly[1], Abdelrahman Salah[2], Hesham M. Eraqi[3]

Date: 10/10/2021

## 1. Abstract:

The largest dataset of Arabic speech mispronunciation errors in Egyptian dialogues is introduced. The dataset is composed of annotated audio files representing the top 100 words that are most frequently used in the Arabic language, pronounced by 100 Egyptian children (aged between 2 and 8 years old). The dataset is collected and annotated on segmental pronunciation errors by expert listeners.

## 2. Dataset Collection:

We collected 100 audio recordings from 100 children in Egypt, where some children pronounce 100 words, and others pronounce only 50 words. The dataset is managed and separated into 100 folders; each contains files of 100 or 50 pronounced words. The dataset can be accessed on Mendeley and Google drive, see [1][4].

## 3. Dataset Description:

The speech is recorded using the Audacity software tool, producing audio files with a Mono channel (one-channel) having a 44.1 kHz sampling rate and a 32-bit resolution. The vocabulary items consist of 100 isolated Arabic words. The database assembly pipeline starts with kids from nursery schools uttering words by pointing at them the words and waiting for them to pronounce the 100 words. Afterwards, the audio files are split into separate audio files, one per word. It is obvious when the voice recording is executed at the school or in the class, there must be noise around or in the recording environment which has a negative effect on speech recognition. We need audacity software to remove the noise. The data is then annotated by labelling every pronounced word file as either correctly or wrongly pronounced. The label information also includes an ID representing the pronounced words from 0 to 99. The dataset is motivated by the lack of data for training or fine-tuning speech representation models, such as the "wav2vec" model [2], and "HuBERT" [3], and to facilitate the development of Arabic language pronunciation mistake identifiers [4],[5],[6].

The dataset is organized as follows. The folder name indicates the child's gender and the number of pronounced words. Folders numbered from 00 to 30 contain audio recordings of children pronouncing 100 words. While folders numbered from 31 to 99 contain audio recordings of children pronouncing 50 words, all files in each folder have index orders from 01 to 50 or from 01 to 100 in cases of pronouncing 50 or 100 words, respectively. If the word is mispronounced by a child, "_N" is added to the word index.

---

[1] CS& Math, Faculty of Science, Fayoum University, Egypt
[2] Nahdat Misr AI, Egypt
[3] CSE & MENG, American University in Cairo (AUC), Egypt.
[4] The dataset link: https://drive.google.com/drive/folders/1dhIp-L0n6_RAzoosVK4bRa7hxBnzebqs

| Index | Word | Index | Word | Index | Word | Index | Word | Index | Word |
|---|---|---|---|---|---|---|---|---|---|
| 1 | نعم | 21 | الطريق | 41 | للغاية | 61 | المدرسة | 81 | ولد |
| 2 | رجل | 22 | عمل | 42 | فتاة | 62 | الصباح | 82 | رسالة |
| 3 | بخير | 23 | الجميع | 43 | كبيرة | 63 | الماء | 83 | عائلة |
| 4 | شخص | 24 | جيدة | 44 | آسفة | 64 | التحدث | 84 | القائد |
| 5 | الوقت | 25 | المال | 45 | الأرض | 65 | الساعة | 85 | المرأة |
| 6 | اليوم | 26 | الذهاب | 46 | البيت | 66 | الليل | 86 | الطبيب |
| 7 | صحيح | 27 | أرجوك | 47 | صباح | 67 | نهاية | 87 | اسم |
| 8 | أستطيع | 28 | المنزل | 48 | ألم | 68 | حياة | 88 | النقود |
| 9 | شكرا | 29 | الحياة | 49 | لحظة | 69 | الواقع | 89 | الكلام |
| 10 | الناس | 30 | انتظر | 50 | بالضبط | 70 | الطفل | 90 | مدينة |
| 11 | أعلم | 31 | الرجال | 51 | رقم | 71 | دكتور | 91 | مساء |
| 12 | رائع | 32 | الله | 52 | طريق | 72 | الهاتف | 92 | الشمس |
| 13 | مرحبا | 33 | الباب | 53 | المدينة | 73 | الطعام | 93 | ارجوك |
| 14 | آسف | 34 | جميل | 54 | الرئيس | 74 | فريق | 94 | السماء |
| 15 | تعال | 35 | الشرطة | 55 | صديقي | 75 | الفتى | 95 | الزواج |
| 16 | بالطبع | 36 | السيارة | 56 | ساعة | 76 | اللقاء | 96 | أصدقاء |
| 17 | العالم | 37 | النار | 57 | غرفة | 77 | نظرة | 97 | مكتب |
| 18 | الحقيقة | 38 | عظيم | 58 | عام | 78 | النساء | 98 | البحر |
| 19 | الليلة | 39 | الخير | 59 | الأطفال | 79 | العشاء | 99 | الكتاب |
| 20 | أمي | 40 | حالك | 60 | سنة | 80 | الأسبوع | 100 | الشارع |

Table 1: 100 most frequently used Arabic words in Egyptian dialogue.

| speaker | Num. of Errors | speaker | Num. of Errors | speaker | Num. of Errors |
|---|---|---|---|---|---|
| 00 | 12 | 10 | 7 | 20 | 15 |
| 01 | 12 | 11 | 15 | 21 | 15 |
| 02 | 7 | 12 | 8 | 22 | 12 |
| 03 | 7 | 13 | 29 | 23 | 15 |
| 04 | 23 | 14 | 9 | 24 | 15 |
| 05 | 6 | 15 | 9 | 25 | 14 |
| 06 | 28 | 16 | 7 | 26 | 6 |
| 07 | 16 | 17 | 6 | 27 | 5 |
| 08 | 24 | 18 | 11 | 28 | 36 |
| 09 | 13 | 19 | 6 | 29 | 11 |

Table 2: number of errors for speakers indexed from 00 to 29

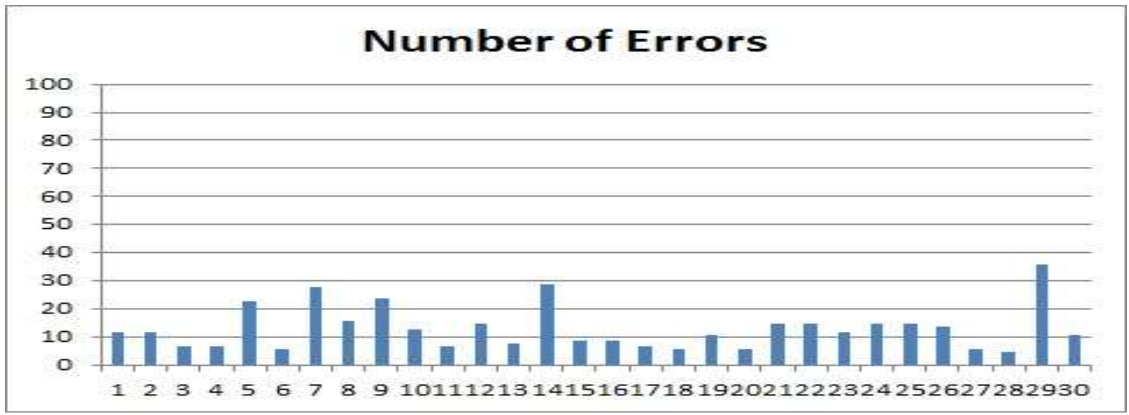

Fig. 1: number of errors for speakers indexed from 00 to 29, and each speaker pronounces 100 words

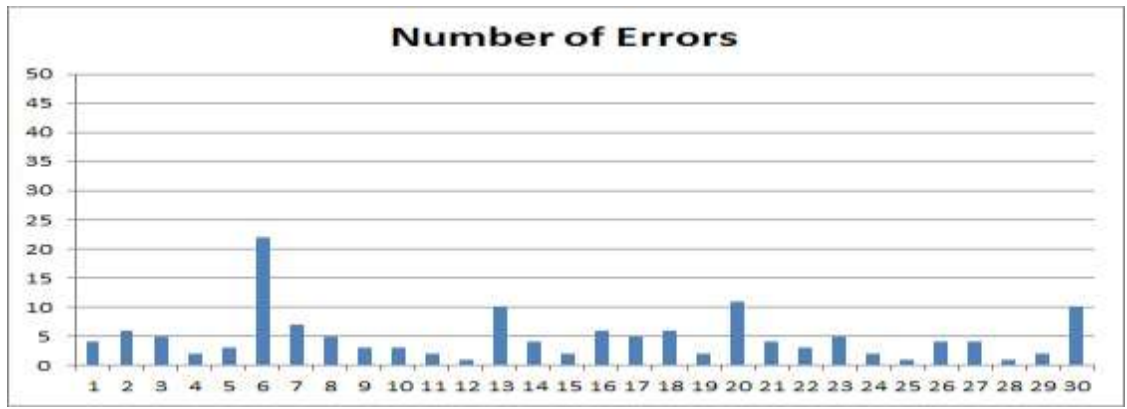

Fig. 2: number of errors for speakers indexed from 31 to 60, and each speaker pronounces 50 words

We organized the dataset into folders and files that contain the 100 audio recordings of 100 children. We used Audacity software [7] to cut and annotate the audio files. We represented an analysis of this dataset as shown in Table 2, and Figures 1 and 2.

## 4. Acknowledgement:
We are thankful to the students and teachers who participated in the dataset collection process.
**Contact:** Please contact salahuqu@gmail.com to obtain a copy of the ASMDD dataset.

## 5. References:

1) Aly, Salah A et. (2021), "Dataset_Arabic_speech_mispronunciation_detection", Mendeley Data, V1, doi: 10.17632/x54dg53rmr.1
2) Baevski, Alexei, et al. "wav2vec 2.0: A framework for self-supervised learning of speech representations." arXiv preprint arXiv:2006.11477 (2020).
3) S. Akhtar, F. Hussain, F. R. Raja, M. E. ul haq, N. K. Baloch, F. Ishmanov, and Y. B. Zikria, "Improving mispronunciation detection of Arabic words for non-native learners using deep convolutional neural network features," Electronics, June 2020
4) D. Korzekwa, J. Lorenzo-Trueba, T. Drugman, S. Calamaro, and B. Kostek, "Weakly-supervised word-level pronunciation error detection in non-native English speech," 2021.
5) D. Korzekwa, R. Barra-Chicote, S. Zaporowski, G. Beringer, J. Lorenzo Trueba, A. Serafinowicz, J. Droppo, T. Drugman, and B. Kostek, "Detection of lexical stress errors in non-native (l2) English with data augmentation and attention," 2021.
6) A. Baevski, M. Auli, and A. Mohamed, "Effectiveness of self-supervised pre-training for speech recognition," arXiv:abs/1911.03912, 2019.
7) Audacity software version 3.1.0 available at: https://www.audacityteam.org/download/windows/